\pdfoutput=1 
\documentclass[10pt, conference, compsocconf]{IEEEtran}
\ifCLASSINFOpdf
\else
\fi

\usepackage{comment}
\usepackage{times}
\usepackage{graphicx}
\usepackage{verbatim}
\usepackage[dvipsnames]{xcolor}
\usepackage{lipsum}
\usepackage{algorithm}
\usepackage{algorithmic}

\usepackage[belowskip=-8pt,aboveskip=1pt]{caption}
\begin{document}

\title{From Knowledge Map to Mind Map: Artificial Imagination}







%
\author{
\IEEEauthorblockN{
Ruixue Liu\IEEEauthorrefmark{1}\IEEEauthorrefmark{4},
Baoyang Chen\IEEEauthorrefmark{2}\IEEEauthorrefmark{4},
Xiaoyu Guo\IEEEauthorrefmark{1}, 
Yan Dai\IEEEauthorrefmark{1},
Meng Chen\IEEEauthorrefmark{1},
Zhijie Qiu\IEEEauthorrefmark{2},
Xiaodong He\IEEEauthorrefmark{3}}

\IEEEauthorblockA{\IEEEauthorrefmark{1}JD AI Platform \& Research, Beijing, China\\}
\IEEEauthorblockA{\IEEEauthorrefmark{2}Faculty of School of Experimental Arts, Central Academy of Fine Arts, Beijing, China\\}
\IEEEauthorblockA{\IEEEauthorrefmark{3}JD AI Research, Beijing, China\\}
\{liuruixue,guoxiaoyu5,daiyan5,chenmeng20,xiaodong.he\}@jd.com, \{chenbaoyang, qiuzhijie\}@cafa.edu.cn
\IEEEauthorblockA{\IEEEauthorrefmark{4}The two authors contributed equally to this work.}
}


\maketitle

\begin{abstract}
Imagination is one of the most important factors which makes an artistic painting unique and impressive. With the rapid development of Artificial Intelligence, more and more researchers try to create painting with AI technology automatically. However, lacking of imagination is still a main problem for AI painting. In this paper, we propose a novel approach to inject rich imagination into a special painting art Mind Map creation. We firstly consider lexical and phonological similarities of seed word, then learn and inherit artist's original painting style, and finally apply Dadaism and impossibility of improvisation principles into painting process. We also design several metrics for imagination evaluation. Experimental results show that our proposed method can increase imagination of painting and also improve its overall quality. 
\end{abstract}

\begin{IEEEkeywords}
Artificial Imagination; Mind Map; AI painting; Dadaism
\end{IEEEkeywords}

%
\IEEEpeerreviewmaketitle

\section{Introduction}
\label{set:intro}
Art and Technology have been interweaved in the course of human history. In Renaissance, the advancement in anatomy led to realistic and perspective depiction of human body. During industrial revolution, the development of photography steered art history away from realism to impressionism and expressionism to capture the beauty of changing nature and inner expression.
Nowadays, Artificial Intelligence (AI) demonstrates stronger potential for art creation. Many researches have been conducted to involve AI into poem generation \cite{c20,c21}, creation of classical or pop music \cite{c27,c28} and automatic images generation \cite{c24,c25,c30}. Whereas there are few researches exploring the possibility of artificial imagination. In this paper, we want to shed more light on if AI has imagination, and we will tackle a more challenging task: the generation of Mind Map with a given topic or idea. 
\begin{figure}[h]
  \centering
  \includegraphics[width=1\linewidth]{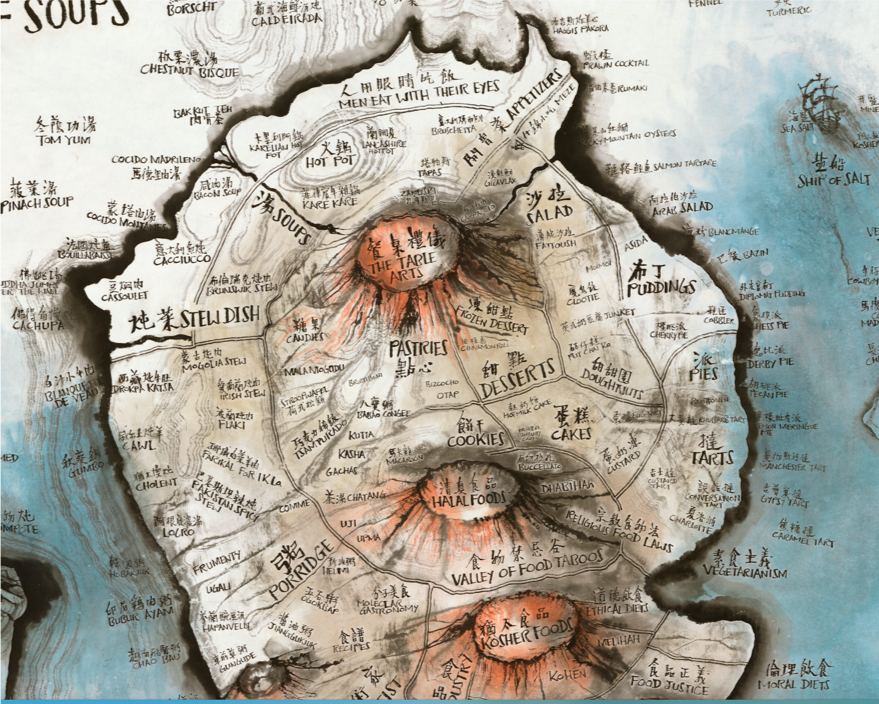}\\
  \caption{An example of Mind Map}
  \label{fig:mindmap}
\end{figure}
Firmly implanted in a scientific epistemology, the \textbf{Map} is viewed as a representation of knowledge. Whereas \textbf{Mind Map} is an artistic representation to visualize the abstract information in a figurative way (Shown in Figure \ref{fig:mindmap}). Sharing the same form with conventional maps, Mind Map is a special diagram which connects and arranges related words, terms and ideas around a central keyword or idea \cite{c16}. Thus, the core issue for Mind Map creation is the information expansion, that is, given a keyword or idea, how to extend it with more ideas, concepts or ontologies. In that case, during this research, instead of discussing the drawing techniques for the visual representation of Mind Map, we focus on the AI-enabled imagination for concepts, ideas and all forms of information expansion in Mind Map.

Currently researches on word expansion, which is mainly based on word embedding techniques \cite{c9}, is a knowledge aggregation rather than imagination.
To generate an imaginative and creative Mind Map, we are facing with the following challenges. First of all, language is an informative and yet complex system for human mind expression. Words and their relations should be extended with many information, such as the lexical, phonologically, psychological features. Thus how to expand imagination beyond the semantic boundary is the first concern in this research. Secondly, same with other artwork, Mind Map needs to reflect individual artist's mind, understanding and experience of the world. And the concept of imitation also laid the historical foundations in both Chinese and Western art \cite{c17,c18}. The second challenge is how to teach AI to learn from artist's mind, knowledge, and experience during the Mind Map creation. Thirdly, unlike cartographers, artists pay less attention on either following the convention or defining accurate relation between concepts. Instead, they plant the seed of artistic free play of imaginations when defining connections between objects or concepts. And the imaginative connections between literal representations and the metaphorical interpretations create possibility to semantic expansion flourished artistically. Our third challenge is how can we transcend from simply  following the convention rules to breaking the restrictions of domain, categorical information and regular convention. 

To solve above challenges, we propose an AI functioned Mind Map generator. Specifically, to better imitate artist's idea and knowledge in AI creation, we propose to establish a knowledge graph by extracting author's ideas and thinking. Further, we break domain and restrictions for word expansion and try to explore more possibilities for words connection by creating rules influenced by the Dadaism\footnote[1]{https://en.wikipedia.org/wiki/Dada}. 

The contribution of this research is three-fold:
\begin{itemize}
  \setlength\itemsep{0em}
    \item We provide a framework to understand how AI can aid in artistic practice. AI in art has more implications than a way of artistic creation, and it enriches the possibility of artistic practice. Further, AI can be considered as co-author, when involves in art making with other artists.
    \item We propose an effective way to inject Artificial Imagination into making artworks in the form of Mind Map, which includes considering semantic similarity, integrating informative linguistic features, inheriting author's mind, and inserting Dadaism and impossibility of improvisation principles.
    \item We situate our work both in arts and the field of Artificial Intelligence, and further develop a set of metrics to evaluate the quality of our proposed methods, which are relevance, linguistic connection, Dadaism, and overall artistic impression.

\end{itemize}

\section{System Overview of Mind Map}

\label{sect:previous}
Mind Map is a special type of drawing in the field of art, combining imagination of concept expansion and visualisation. Whereas the essential idea about Mind Map is the expansion of concepts or ideas related to a keyword, which this research is focused on.
\subsection{System Overview:} 
\begin{algorithm}[h] 
\caption{ Mind Map System} 
\label{alg:Framwork} 
\begin{algorithmic}[1] 
\REQUIRE ~~\\ 
seed words the user entered, $W$; \\
\ENSURE ~~\\ 
A Mind Map image, $M$;
\FOR{each $w \in W$}
\STATE Drawing the seed word $w$ on the Mind Map image $M$
\STATE Expanding the seed word $w$ to a list of candidates $w_1, w_2, ... w_n$
\FOR{each $w' \in [w_1, w_2...w_n]$}
\STATE Classifying the domain of the word $w'$
\STATE Mapping a painting element to the current word $w'$
\STATE Calculating the distance from the candidate word $w'$ to the seed word $w$
\STATE Drawing the word $w'$ on the Mind Map image $M$
\ENDFOR
\ENDFOR
\RETURN $M$; 
\end{algorithmic}
\end{algorithm}

From Algorithm \ref{alg:Framwork} above, we firstly expand the given seed word into a group of candidates based on a set of criteria. Then, we classify the candidate word into one of 6 domains, which are architecture, mountain, river, grassland, road and lake. These domains are summarised from the representative Mind Map paintings from an artist. With the domain information, we can project the invisible word meaning into specific visible painting element, which constructs the basic concept/object in the Mind Map. We also define the distance among words in Mind Map and visualise it as path in pictures. The whole creation process is iterative. At each iteration, the extended candidate can be selected as the seed word for the next round of expansion. The procedure repeats until the final Mind Map picture is presented.

\subsection{Baseline Method}
In the domain of Natural Language Processing, semantic expansion is realised by calculating semantic similarity between words. Studies on semantic similarity matching or semantic expansion include knowledge-based methods \cite{c3,c4} or corpus-driven approaches \cite{c9}. Knowledge-based approaches rely heavily on the existing ontology or taxonomy, and usually can't scale up, while corpus-driven methods learn vector space representations for words from text data, which can capture the semantic and syntactic information. 



\begin{figure*}[h]
  \centering
  \vspace{-3pt}  
  \setlength{\abovecaptionskip}{-3pt}   
  \setlength{\belowcaptionskip}{-20pt}   
  \includegraphics[width=0.9\textwidth]{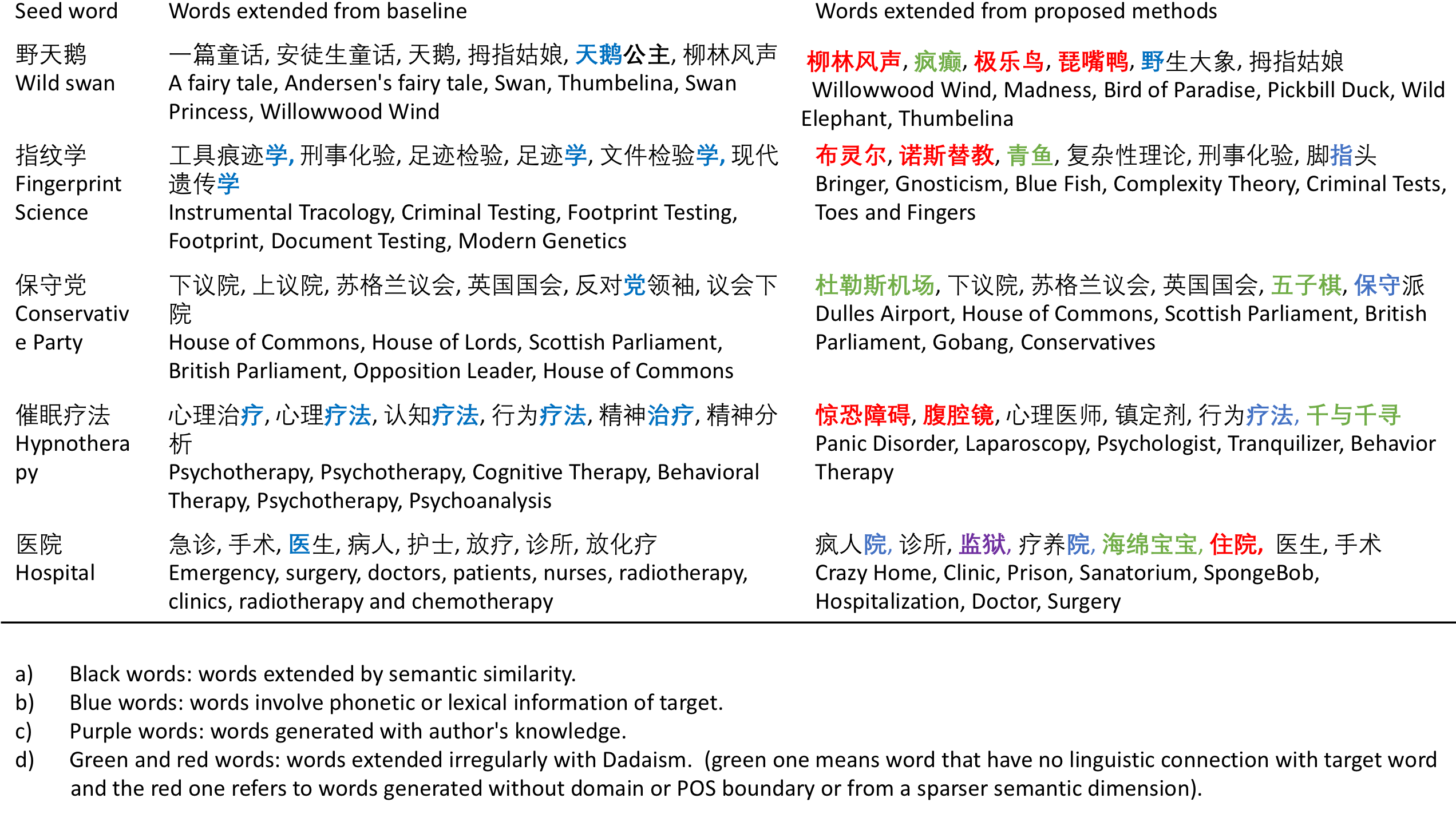}
  \caption{Examples of word expansion from baseline and proposed methods.}
  \label{fig:qunt_result}
\end{figure*}

\label{sect:method}
\section{Proposed Method}\label{set:methods}
In this research, we aim to create a Mind Map generator which can not only expand word based on semantic meanings but also reflect the creative and imaginative nature of art work. The corpus-driven semantic expansion is the basic source for Mind Map generator. To increase the diversity and creativity, we firstly consider informative linguistic features, then learn and inherit artist's original mind and style, and finally bring Dadaism into the process. We will show that the proposed methods can increase imagination of AI painting significantly.

\subsection{Integrate Informative Linguistic Features}
Previous studies show that linguistic feature is very important in children's language acquisition and is also the intrinsic connection of words. According to \cite{c11}, there are five designed features of language: arbitrariness, duality, creativity or productivity, displacement and cultural transmission, where duality refers to two levels of language structure (primary level for meaningful word units and the secondary level for meaningless sounds). This reveals that lexical and phonetic features contribute to the diverse meaning expression for language.

To address linguistic features in Mind Map, we adopt the morphological and phonological information into word expansion. We propose a rule-based method of examining words that share similar characters or homophonous syllables\footnote[2]{Homophonous syllables here refer to the relation between different lexical items which have unrelated meanings but accidentally exhibit an identical phonetic form} of seed word. As Chinese is rich in ambiguous words, its characters or words are full of connotations and associations. Exploring the polysemy possibility of Chinese characters/words with the lexical items will also extend the semantic diversity for the seed word and can bring unexpected imagination. Moreover, due to the limited phonetic inventory, Chinese has a significant number of homophonic syllables or words, which are frequently used in Chinese poetry or puns. The investigation of phonetic information in seed word expansion will jump the traditional semantic boundary in Mind Map.

For lexical features, we adopt the longest sub-common string to examine the candidate Chinese words for the seed word. While for phonological information, we firstly convert each seed word into its phonetic representation and then go through the vocabulary to find the words sharing the same phonetic syllable.

\subsection{Learn and Inherit Artist's Mind}
During creating artworks, both eastern and western thinkers have both vastly invested in the importance of imitation. In history, the Chinese painterly tradition of imitation (Transmission by Copying\cite{c17}) shares the similar notion. The Shan Shui painter repeatedly imitated the painting of the genre before them which eventually introduced subtle variations and led to progress in the large time scale.Based on spiritual reflections of ideas, artists form their own distinct imaginative minds imperceptibly.
Therefore, to improve the imagination of AI Mind Map generation, we propose a mind adaptation method that takes artist's paintings as the source of creative inspiration to produce specific artworks.

Specifically, we develop a knowledge graph to imitate and reproduce respective artist's creative and imaginative work. Consisting of words and relations among them, the tree structure graph treats words as vertexes and relations as edges. Note that relations are either hyponym or hypernym in this structure. To construct the knowledge graph, firstly, we obtain inherent and representative words extracted directly from artist's creations. Secondly, according to instructions from these words, we define some topic words that are most relevant to primitive words and prepare further expansion. Finally, based on the artist's understanding of certain topics and domains, we convert seemly disorder imagination of artists into ordered word expansion rules. What's more, we construct relations between words to imitate thinking procedures of artists. The reason why we employ the non-linear thinking of the artist is because these features can imitate human creation and imagination to the utmost extent.

\subsection{Integrate Dadaism Principle}
When representing the terrain of the mapped object on flat media, there are always gaps between linked objects shown in map. Cartographers try to compensate those inaccuracies via different mapping techniques by preserving different metric properties. However, artists often break the lines between the objects shown in a map, which offers an alternative for artists when creating and understanding maps of all kind. They pay less attention in following the convention and making the map accurately, but plant the seed of free play of imaginations in those gaps. In that case, an outstanding feature for Mind Map development is breaking rules of convention and exploring possibilities for unexpected connections between ontologies. 

In order to integrate these features into Artificial Imagination, we seek to follow above theories in artwork creation. Dadaism is an important art movement in history which rejects the logic, reason, and encourages to express irrationality and anti-bourgeois in the art work. One of the most typical Dadaist works from Marcel Duchamp is the \textit{Fountain}, in which he broke the semantic boundary between \textit{plumbing} and \textit{fountain} and offered an surprising metaphor. 

Following the internal feature of \textit{Dada Poem Generator} and Dadaist artworks, we have developed three principles for seed word expansion in Mind Map (although the only rule for Dada is \textit{Never follow any known rules}):
\begin{itemize}
  \setlength\itemsep{0em}
    \item Jumping the domain boundaries for word expansion by exploring cross-domain vocabularies;
    \item Enriching semantic diversities by exploring different part-of-speech tagging words from seed word;
    \item Breaking the linguistic restrictions by investigating random candidates beyond semantic, lexical or phonetic connections. 
\end{itemize}
\begin{figure*}[h]
  \centering
  \includegraphics[width=1\textwidth]{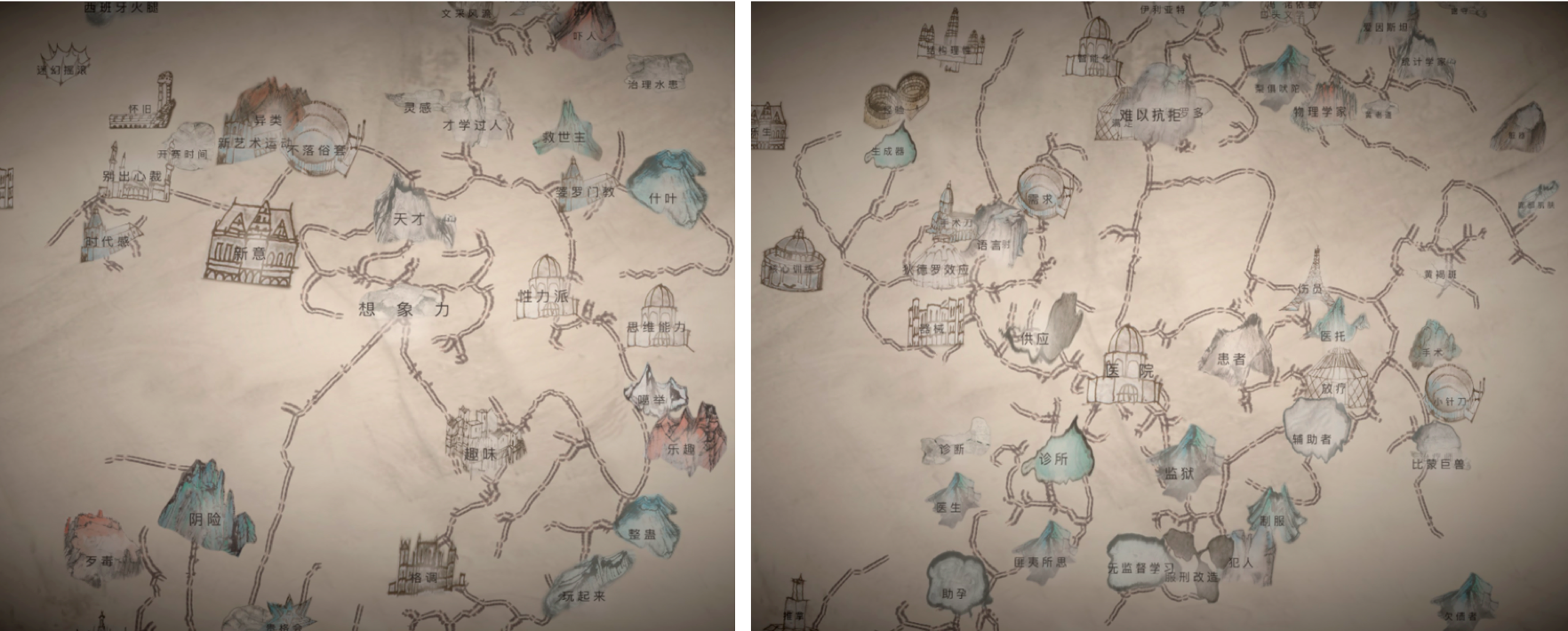}
  \caption{Examples of generated Mind Maps with topic of Imagination (left) and Hospital (right)}
  \label{fig:maps}
\end{figure*}

\section{Experiment}\label{set:exper}
\subsection{Experiment Setup}
Instead of training our own word embedding from large corpus, we use the open resource word embedding for Chinese words and phrases \cite{c13}. The available corpus provides 200-dimension vectors representations and covers 8 million Chinese words and phrases \footnote[3]{https://ai.tencent.com/ailab/nlp/embedding.html}. The baseline methods for semantic word expansion is based on word vectors and calculated with cosine similarity.

To facilitate the development of Mind Map, we collect ten Mind Map artworks from one of the famous artists in this field (the same artist from whom we extract the painting elements). Each of the selected map is created within a certain domain or topic, including food, occupations, clothes, AI, body, stories and fairy tales, sports, spiritual feelings and religious etc. 

We then extracted around 5,000 unique words or expressions from these maps. We filtered the irregular expression for those more than eight Chinese characters and those can not be found in the word embeddings. Later we took these words as seed word, and extended the number into around 300, 000 candidates with the cosine similarity method. To maintain the high quality of word expansion, we also invited ten university students majored in \textit{Experimental Art} to manually examine these words and filter the irrelevant ones for certain domains. After manual examination, around 15,500 words are used for candidates in Mind Map generation.

Moreover, with the additional knowledge expansion provided by the artist, we created a knowledge network including 1754 relations and 12,000 entities. This knowledge network is later integrated into Mind Map development as mentioned in Section \ref{set:methods}. 

For human evaluation, we extracted 130 words from the corpus and then extend them with baseline and the proposed methods. For each seed word, we remain at most 7 candidates for comparison and further analysis.

\subsection{Evaluation Metrics}\label{sec:human_eval}
The evaluation of extended words or expressions is generally a challenging task and there are no established metrics in previous works, not mentioning the words generated for artwork creation. To better address the performance of the proposed method, we conduct extensive studies in three ways: manual evaluation, quantitative analysis and qualitative analysis methods.

We invited 8 experts to participate in the experiment for human evaluation, who have rich experience for art creation. We also design four metrics for human evaluation as follows:

\textbf{Relevance:} it measures whether the expanded words are related to the seed word. Words that are more similar to the seed word will gain higher scores.

\textbf{Linguistic Connection:} it reflects the lexical and phonological information of words when compared with the seed word. For words in the same sub-common string or sharing  more similar phonetic syllable with the seed word, they gain a higher score.

\textbf{Dadaism:} it represents a criterion for Dadaism of the expansion results. The stronger Dadaism words have, the higher score they obtain.

\textbf{Overall Impression:} it reflects the general impression, creativity and imagination of words. A higher score means that it leaves a more striking impression on the viewers.

\begin{table}
\caption{Word expansion distributions}
\centering
\begin{tabular}{|l|l|l|l|l|}
\hline
\begin{tabular}[c]{@{}l@{}}\end{tabular} & \begin{tabular}[c]{@{}l@{}}Semantic \\ Similarity\end{tabular} & \begin{tabular}[c]{@{}l@{}}Linguistic\\  Feature\end{tabular} & Dadaism & \begin{tabular}[c]{@{}l@{}}Author\\
Style\end{tabular} \\ \hline
Baseline & 67.36\% & 25.82\% & 4.40\% & 2.42\% \\ \hline
\begin{tabular}[c]{@{}l@{}}Proposed\\ Method\end{tabular} & 35.16\% & 26.37\% & 22.86\% & 15.61\% \\ \hline
\end{tabular}
\label{table:distrib}
\end{table}

In addition to human evaluation, we also invite three of the experts to manually annotate the method used for each word generation and accepted their annotation when two or more people share the same opinion. In total, four types of expansion method are involved: 1) semantic similarity, 2) linguistic feature, 3) author style and 4) Dadaist method. The detail analysis on distribution of used methods are demonstrated in the section below.

\subsection{Results \& Analysis}
\subsubsection{Qualitatively Analysis}

We firstly compared the percentage of different types of word expansion between baseline and proposed approach. To calculate this, we classify each extended word candidate into 4 categories. Semantic Similarity means the candidate word is from word vector similarity. Linguistic Feature means it is expanded based on either lexical or phonetic rule. Dadaism represents the word is generated based on Dadaism principle. And Author Style means the word exists in artist's original artworks. As shown in Table \ref{table:distrib}, our proposed method outperforms baseline in measurements including \textit{Linguistic information}, \textit{Dada} and \textit{Author mind}. For the latter two criteria, our method obtains far more distributions of words than baseline, which denotes that the proposed method could generate more unexpected (Dadaism) and specific stylized (Author mind) words. 

Note that baseline method gains higher distribution of words in \textit{Semantic similarity} and basically the same percentage in \textit{Linguistic feature}. Higher distribution in \textit{Semantic similarity} is because all of candidate words in baseline are extended based on word vectors, while proposed method has more expansion choices. Furthermore, word vectors can also catch linguistic features sometimes, that's why percentages of \textit{Linguistic feature} are very close between baseline and proposed approach.

\subsubsection{Human Evaluation Results} 
\begin{figure}[h]
  \centering
  \vspace{-8pt}  
  \includegraphics[width=1.0\linewidth]{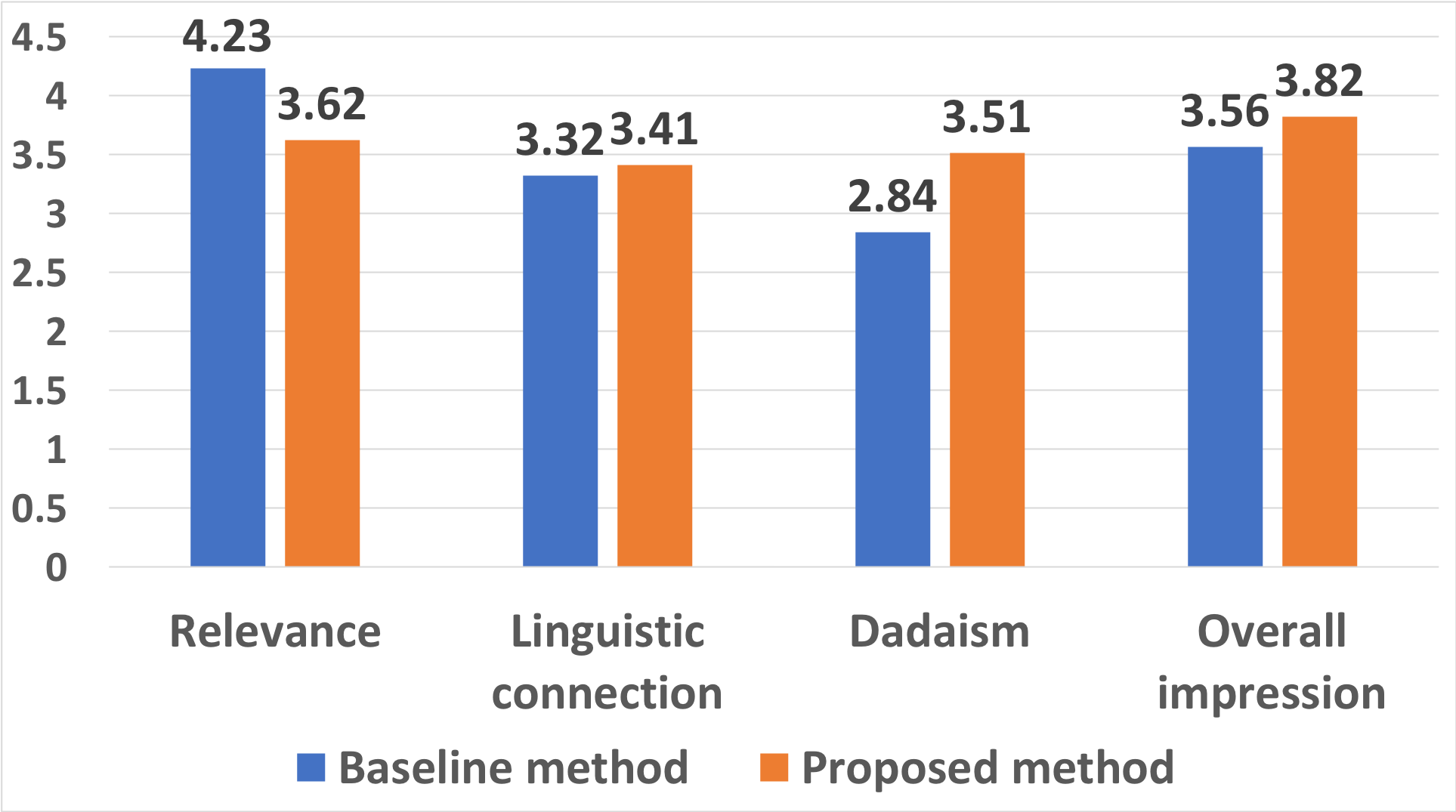}\\
  \caption{Human evaluation results}
   \setlength{\abovecaptionskip}{0.cm}
   \setlength{\belowcaptionskip}{-0.cm}
  \label{fig:human_score}
\end{figure}

The comparison of human evaluation results are shown in Figure \ref{fig:human_score}. Figure \ref{fig:human_score} indicates the average score on the four criteria mentioned in Section \ref{sec:human_eval}. The proposed method outperforms the baseline in the three metrics including \textit{Linguistic Connection}, \textit{Dadaism} and \textit{Overall Impression}. As observed from the first criteria of semantic relevance, the baseline method gets higher score than the proposed methods. This is because only semantic similarities is concerned in baseline. For the second criteria, we observe that, the experts have a little bit higher preference on the informative linguistic features involved in word expansion by the proposed methods. However, for Dadaism, the proposed methods generated many more unexpected and imaginative candidate words. Finally, the linguistic informative features and Dadaism from the proposed methods contribute together to the overall expression of Mind Map generation, which won expert's preference for the overall impression.

\subsubsection{Analysis and Discussion}
\label{set:quanti}
Figure \ref{fig:qunt_result} shows the examples of words generated from the baseline method and the proposed method given the same seed word. To interpret the results in details, we marked words generated with different methods in various colors. In general, our proposed methods cover all kinds of expansion strategies and more imaginative. For example, in first row, we can see that \textit{Madness} is extended from \textit{Wild swan}, however it's hard to find explicit relationship between these two words. But when these two words are connected together, it describes a picture of how the elegant swan are disturbed by hunters and fly away in a mad way. So this is the result of considering Dadaism. In the fifth row, with lexical information, we successfully extend the noun word \textit{hospital} to the verb \textit{hospitalization} (hospitalization is a verb in Chinese). More interesting, \textit{prison} is extended from \textit{hospital} too, which actually inherits from artist's original mind. The artist believes that \textit{hospital} is \textit{prison} where patients lose their freedom of health inside. All these examples show that our proposed approach brings more imagination into word expansion.  

In addition, our method can also increase the diversity of graphics and enrich the aesthetic pleasure for the audience. Figure \ref{fig:maps} are two paintings created by our system, which are the final results of 28 generation iterations. We can see that they are generated with Chinese Shan Shui painting style. The topic of the left picture is imagination, and the right has the topic of hospital. The program surprisingly paired a lake with our input word imagination, which poetically gives the viewer a sense of thoughts springing out of it. This indicates that by injecting imagination into AI painting, the artistic quality of paintings can be improved too.

\section{Conclusions}
As the frontier work of AI supported imagination exploration, we propose a novel approach to increase the creativity, linguistic diversity and irregularity in Mind Map creation. By combining semantic similarity, linguistic connection, artist's original mind and Dadaism, we improve the imagination of AI painting significantly. We also designed four novel evaluation metrics to measure imagination. Experimental results and extensive analysis show the effectiveness our proposed methods with respect to \textit{Artificial Imagination}. In the future, we plan to inject imagination into the visualisation process of Mind Map creation, and explore more painting categories too.

\section*{Acknowledgment}
This paper stems from a collaboration between JD AI Research and Central Academy of Fine Arts (CAFA) in Beijing, China. We are excited to be greeted by our time that exploring creativity in AI is both intriguing and achievable, yet more to endeavor in the field of art as science and science as art. We want to thank both institutions for their support, and additionally CAFA EAST AI Group.




{
\small
\bibliographystyle{IEEEtran}
\addtolength{\itemsep}{-1.5ex}
\bibliography{main}}

\end{document}